\documentclass[10pt,twocolumn,letterpaper]{article}
\usepackage[pagenumbers]{cvpr} 

\definecolor{cvprblue}{rgb}{0.21,0.49,0.74}
\usepackage[pagebackref,breaklinks,colorlinks,allcolors=cvprblue]{hyperref}

\usepackage{graphicx}
\usepackage{amsmath}
\usepackage{amssymb}
\usepackage{booktabs}
\usepackage{enumitem} 
\usepackage{setspace}  
\usepackage{booktabs} 
\usepackage{listings} 
\usepackage{xcolor} 
\usepackage{enumitem}
\usepackage{subcaption}
\usepackage{graphicx}
\usepackage{multirow}
\usepackage{graphicx} 
\usepackage{float}    
\usepackage{adjustbox} 
\usepackage{amssymb}  
\usepackage{graphicx}
\usepackage{pifont}
\usepackage[hang,flushmargin]{footmisc}

\def\paperID{15808}
\def\confName{CVPR}
\def\confYear{2025}

\title{Seeing Soundscapes: Audio-Visual Generation and Separation\\from Soundscapes Using Audio-Visual Separator}

\author{
    Minjae Kang \qquad Martim Brandão\\
    King's College London, United Kingdom\\
    \texttt{\small \{minjae.kang, martim.brandao\}@kcl.ac.uk}
}

\begin{document}

\twocolumn[{
\renewcommand\twocolumn[1][]{#1}
\maketitle
\begin{center}
    \centering
    \captionsetup{type=figure}
    \includegraphics[width=\linewidth]{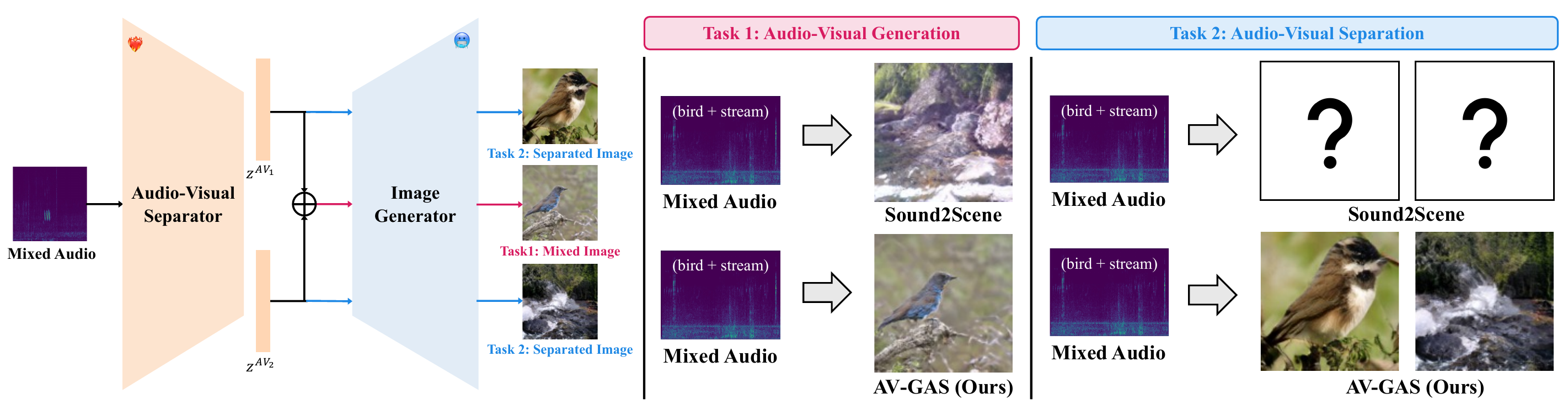}
    \captionof{figure}{\textbf{Comparison between our approach and existing methods.} Our approach processes a mixed audio input to generate images, whereas existing methods generate images given single-class audio and fail to generate plausible images given mixed audio. Our method can be used for two tasks: first, to generate an image containing all classes present in the audio~(task1: audio-visual generation~(coloured in red)); second, to generate a separate image for each object present in the audio~(task2: audio-visual separation~(coloured in blue)). Our model is the first that can generate single images or multiple class-separated images from mixed audio.} 
    \label{fig:fig1}
\end{center}
}]
\begin{abstract}
{
    Recent audio-visual generative models have made substantial progress in generating images from audio. However, existing approaches focus on generating images from single-class audio and fail to generate images from mixed audio. To address this, we propose an Audio-Visual Generation and Separation model (AV-GAS) for generating images from soundscapes (mixed audio containing multiple classes)~\footnote{Originally submitted to CVPR 2025 on 2024-11-15 with paper ID 15808.}. Our contribution is threefold: First, we propose a new challenge in the audio-visual generation task, which is to generate an image given a multi-class audio input, and we propose a method that solves this task using an audio-visual separator. Second, we introduce a new audio-visual separation task, which involves generating separate images for each class present in a mixed audio input. Lastly, we propose new evaluation metrics for the audio-visual generation task: Class Representation Score~(CRS) and a modified R@K. Our model is trained and evaluated on the VGGSound dataset. We show that our method outperforms the state-of-the-art, achieving 7\% higher CRS and 4\% higher R@2\textsuperscript{*} in generating plausible images with mixed audio.
    }
\end{abstract}
\vspace{-\baselineskip}

\section{Introduction}
\label{sec:intro}
    Audio-visual generation aims to generate visuals from audio signals or vice versa~\cite{sung2023sound}. Although generating plausible visuals or audio signals from the other modality is challenging due to the differences in representations and the gaps between these two modalities (audio and visual)~\cite{sung2023sound}, there have been multiple attempts at solving this challenge. In particular, audio-to-image generative models have successfully achieved to generate plausible images from single-class audio input~\cite{sung2023sound, av_vae_pedersoli2022estimating(Baseline), av_gan7_fanzeres2022soundtoimagination(S2I)}.
    
    Recent approaches for the audio-to-image generation task focus on generating images from a single-class audio input~(e.g., bird calling). However, we are surrounded by multiple audio sources~(e.g., bird calling and stream burbling), and these approaches fail to generate images containing two or more objects when input audio includes various classes, as we will show. This limitation of previous approaches leads us to the research question: \textit{``Can machines imagine a scene from soundscapes~(mixed-class audio)?"}.

    To this end, we propose \textsc{AV-GAS}: an \textbf{A}udio-\textbf{V}isual \textbf{G}eneration \textbf{A}nd \textbf{S}eparation model that generates images from mixed audio. As shown in Fig.~\ref{fig:fig1}, our proposed model can be used in two ways: first, to generate a single image containing all the classes present in the audio~(audio-visual generation); second, to generate a separate image for each class present in the audio~(audio-visual separation). The second use of our method we call ``audio-visual separation'', and we propose it as a new task in audio-visual separation. 

    Since existing evaluation metrics do not capture the goal of image generation tasks from mixed audio, we introduce a new evaluation metric, \textit{Class Representation Score~(CRS)}, and modify R@K for our task given mixed audio. We train our model on audio-visual training samples from VGGSound~\cite{Chen20VggSound} and evaluate it on mixed audio. We show that our approach achieves a 7\% higher CRS and 4\% higher R@2\textsuperscript{*} in generated images.
    
    In summary, our core contributions are as follows:\vspace{-0.1\baselineskip}
    \begin{itemize}[noitemsep, leftmargin=*]
    \item We propose a new audio-visual generative model focusing on generating images from mixed audio for the first time to address the limitation of existing approaches.
    \item We introduce a new audio-visual separation task of generating separated images for each class present in mixed audio.
    \item We introduce audio-visual training input samples for audio-visual generation from mixed audio.
    \item We investigate how the audio-visual generation task given mixed audio can be evaluated by introducing a new evaluation metric called Class Representation Score~(CRS) and modifying the existing R@K metric.
    \end{itemize}

\section{Related Work}
\label{sec:related-work}
\noindent\textbf{Audio-Visual Generation.}
    Audio-visual generation is the task of generating audio signals from visuals or vice versa~\cite{sung2023sound}, categorising the task into audio-to-visual~(A2V) and visual-to-audio~(V2A). A2V is further divided into two tasks: audio-to-image~(A2I) generation and audio-to-video~(A2V) generation. This work focuses on the A2I problem where the goal is to obtain images from audio. While many previous studies have addressed this task, these studies mainly focus on generating images from a single-class audio input, rather than a mixed audio signal~\cite{av_vae_pedersoli2022estimating(Baseline), av_gan1, av_gan2_hao2017cmcgan, av_gan3_Qiu2018ImageGA, av_gan4_Wan2018TowardsAT, av_gan5_9412890, av_gan6, av_gan7_fanzeres2022soundtoimagination(S2I), speech2face_Oh2019Speech2FaceLT, transformer1, contrastive, sung2023sound, yariv2023audiotoken, biner2024sonicdiffusion}. 
    
    Due to the differences between the two modalities, generating images from audio is a very challenging problem~\cite{sung2023sound}. With the rise and development of text-to-image~(T2I) generation models~\cite{pmlr-v139-ramesh21a-dalle, Ramesh2022-dalle2, rombach2021highresolution-stablediffusion}, particularly with the development of diffusion models~\cite{diffusion, DDPM, DDIM, rombach2021highresolution-stablediffusion}, some works have tried to use text information for the A2I generation task by using text tokens to feed into the model or align with audio tokens~\cite{yariv2023audiotoken, biner2024sonicdiffusion} with pre-trained T2I diffusion models~\cite{rombach2021highresolution-stablediffusion}. However, using another modality can introduce unintended biases instead of using information from audio directly. 
    
    On the other hand, other works have tried to address the problem without using text information. These approaches can be categorised by their architectures and learning methods. Some works use Variational Autoencoder (VAE) architectures~\cite{av_vae_pedersoli2022estimating(Baseline)}. After Generative Adversarial Networks (GANs)~\cite{gans} significantly succeeded in various tasks, most of the previous works adopted GAN-based methods~\cite{av_gan1, av_gan2_hao2017cmcgan, av_gan3_Qiu2018ImageGA, av_gan4_Wan2018TowardsAT, av_gan5_9412890, av_gan6, av_gan7_fanzeres2022soundtoimagination(S2I)}, which use the generator and discriminator architectures, and an adversarial loss. More recently, some works have adopted transformer architectures~\cite{transformer1, trasnformer2}, and it is also noteworthy that contrastive learning~\cite{contrastive} has been used to align audio and image embeddings without labels in some works~\cite{sung2023sound}. However, these approaches have limitations in generating images given mixed audio. To address this, we aim to build an audio-to-image generative model to generate images from a mixed audio input.

    \noindent\textbf{Audio-Visual Separation.} 
    We propose a new task in audio-visual separation. This task aims to obtain separate images for each class present in an audio signal. For example, if the audio contains both bird and stream sounds, the goal of the task is to generate two separate images: a bird image and a stream image. There have been separation tasks in audio-visual separation, such as detecting instrument sounds separately from a video~\cite{chen2023iquery, tan2023language} or identifying which speaker is talking in a video~\cite{cheng2023filterrecovery, montesinos2022vovit}. Like sound separation, most of the separation tasks in audio-visual learning focus on speech~\cite{cheng2023filterrecovery, montesinos2022vovit} and music data~\cite{chen2023iquery} rather than general sounds~\cite{tzinis2022avscopev2, chatt2024scenegraph}. Our proposed new task in audio-visual separation can be applied to any type of domain, including speech, music and general sounds. We anticipate that our proposed audio-visual separation task can be applied to various applications, particularly in education and healthcare, by separating audio sources and generating multiple visuals separately.
    
\noindent\textbf{Generative Models.}
    Generative Adversarial Networks (GANs)~\cite{gans} are generative models trained using a generator and discriminator architecture with adversarial loss. In the audio-visual generation task, there are two main approaches when using GANs: using GAN-based training or using a GAN generator for generating images. Some existing works adopt GANs and train their models with adversarial loss~\cite{av_gan1, av_gan2_hao2017cmcgan, av_gan3_Qiu2018ImageGA, av_gan4_Wan2018TowardsAT, av_gan5_9412890, av_gan6, av_gan7_fanzeres2022soundtoimagination(S2I)}. On the other hand, some works train GANs only for an image generator~\cite{sung2023sound}.  
    Diffusion models are another type of powerful generative model. In particular, the development of Large Language Models (LLMs) has impacted the advancement of Text-to-Image (T2I) and Image-to-Text (I2T) generation, enabling diffusion models to be applied to various tasks with text. For the audio-visual generation task, most works use pre-trained diffusion models~\cite{rombach2021highresolution-stablediffusion, biner2024sonicdiffusion} and utilize text information as prompts or during training, which does not align with our goal of using only audio information. 
\section{Method}
\label{sec:method}

\subsection{Audio-Visual Training Input}\label{section:data}
\begin{figure}[h!]
        \centering
        \includegraphics[width=\linewidth]{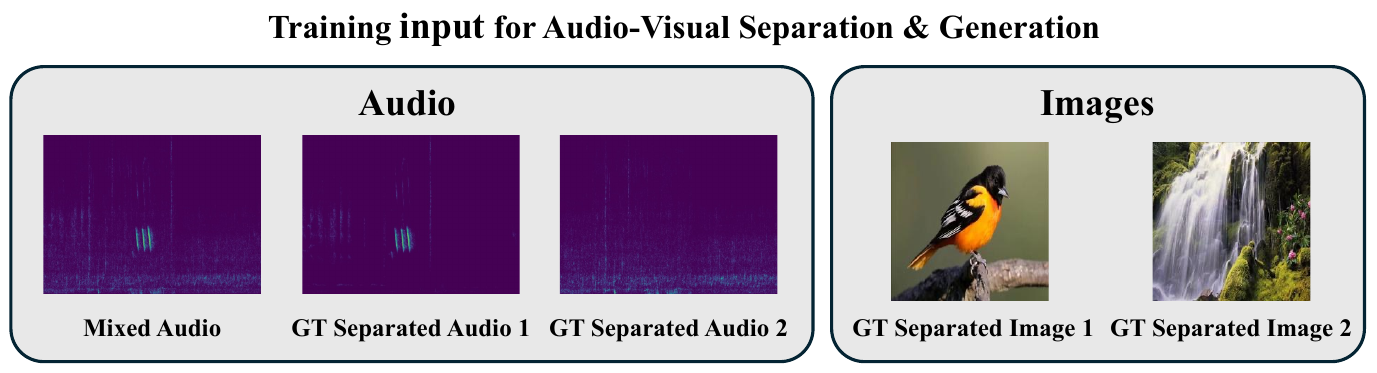}
        \caption{\textbf{Our audio-visual training input example.} We create new audio-visual training input tuples using VGGSound~\cite{Chen20VggSound} to process a mixed audio signal. Since our task aims to generate images from mixed audio, we mix two audio classes and create the audio-visual training input tuples with the mixed audio, two ground truth separated audio for sound separation and two ground truth separated images that cannot be mixed like audio.}
        \label{fig:pairs}
    \end{figure}

    During training, our method takes as input a dataset containing pairs of single-class images and their corresponding sounds (e.g., VGGSound~\cite{Chen20VggSound}). Since audio signals can be easily added together, we generate N combinations of different classes (e.g. `bird' and `stream' is one combination of two classes), and for each combination, we obtain a tuple of: a mixed-audio signal (obtained by addition of the two audio signals), the two original separated audio signals (one for each class), and the two original images (one for each class). Figure~\ref{fig:pairs} shows an example of a training tuple. Our goal is to design a model that, after training on this data, can eventually imagine a scene from a soundscape by first imagining each object individually and then the entire scene.

\subsection{Architecture}
    \begin{figure*}[h]
        \centering
        \includegraphics[width=\linewidth]{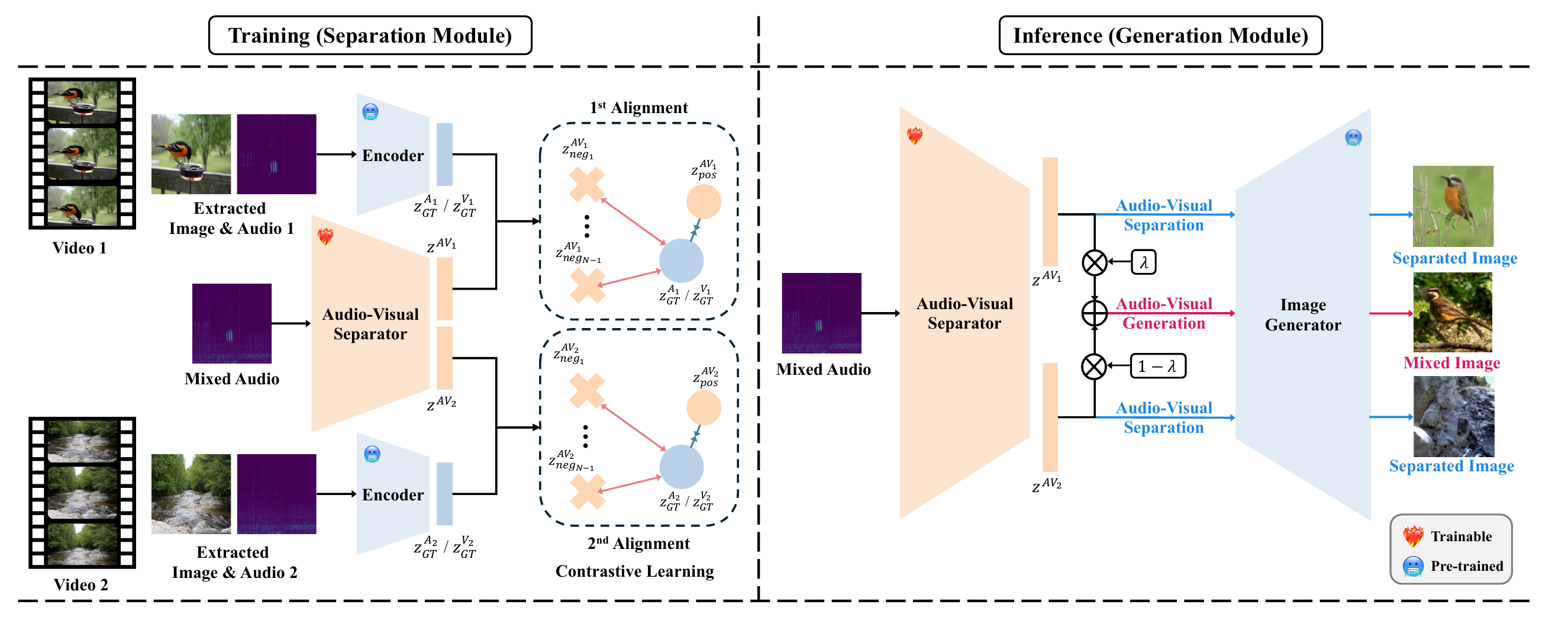}
        \caption{\textbf{Overview of our approach.} Our approach includes two modules: the separation module~(for training) and the generation module~(for inference). In the separation module, audio-visual training input tuples extracted from video samples are used for training. The audio-visual separator learns to distinguish objects in mixed audio using contrastive learning loss without class labels. The audio and/or image embeddings extracted from the pre-trained encoder are compared to the embeddings from the audio-visual separator. In the generation module, the two embeddings extracted from the audio-visual separator are used for the two types of image generation. Each embedding itself generates separate images~(task: audio-visual separation~(coloured in blue)) with the image generator, and if two embeddings are combined with the control parameter $\lambda$, mixed images are generated~(task2: audio-visual generation~(coloured in red)).}
        \label{fig:model}
    \end{figure*}

Our model broadly consists of two modules: a separation module~(an audio-visual separator) and a generation module~(an image generator) as shown in Fig.~\ref{fig:model}. The audio-visual separator is a module for separating audio sources and converting them to embeddings in the audio-visual representation space. The image generator is used for generating an image containing all classes present in mixed audio audio or multiple images each representing a different class. 

\noindent\textbf{Audio-Visual Separator (separation module).} Our audio-visual separator $f_{AV}(\cdot)$, separates audio input into two embeddings as shown in Fig.~\ref{fig:model}. It is similar to the audio encoder as it extracts audio features from the audio input. However, we call it an audio-visual separator because it is aligned with audio and image embeddings during training in the latent space, and the extracted embeddings are used in the generation of images, particularly for our proposed `audio-visual' separation task. We refer to the features extracted by the audio-visual separator as `audio-visual features' to distinguish them from audio features and image features.

We use ResNet-18~\cite{He2015DeepRL-resnet} architecture for the audio-visual separator, which is the same backbone as the baseline's audio encoder~\cite{sung2023sound}, since we want to control for variables that might arise from changes in the architecture of the encoder. The output of the audio-visual separator $f_{AV}(\cdot)$ is a single 4096-dim vector $\mathbf{z^{\scriptscriptstyle{AV_{mix}}}}$. The first half of the embedding vector $\mathbf{z^{\scriptscriptstyle{AV_{1}}}}$ corresponds to one of the classes in the mixed audio, and the second half $\mathbf{z^{\scriptscriptstyle{AV_{2}}}}$ corresponds to the other class.

After the separation, these two separated embedding vectors are compared with the ground truth audio and image embeddings ($\mathbf{z^{\scriptscriptstyle{A_1}}_{\scriptscriptstyle{GT}}}$, $\mathbf{z^{\scriptscriptstyle{A_2}}_{\scriptscriptstyle{GT}}}$, $\mathbf{z^{\scriptscriptstyle{V_1}}_{\scriptscriptstyle{GT}}}$, $\mathbf{z^{\scriptscriptstyle{V_2}}_{\scriptscriptstyle{GT}}}$) extracted by the pre-trained audio encoder $f_{A}(\cdot)$ and the pre-trained image encoder $f_{V}(\cdot)$. The first half of the embedding $\mathbf{z^{\scriptscriptstyle{AV_{1}}}}$ is compared to the ground truth audio and image of one of the classes present in the mixed audio $\mathbf{z^{\scriptscriptstyle{A_1}}_{\scriptscriptstyle{GT}}}$, $\mathbf{z^{\scriptscriptstyle{V_1}}_{\scriptscriptstyle{GT}}}$. Likewise, the second half of the embedding $\mathbf{z^{\scriptscriptstyle{AV_{2}}}}$ is compared to the ground truth audio and image of the other class present in the mixed audio $\mathbf{z^{\scriptscriptstyle{A_2}}_{\scriptscriptstyle{GT}}}$, $\mathbf{z^{\scriptscriptstyle{V_2}}_{\scriptscriptstyle{GT}}}$. We use the pre-trained audio encoder and the pre-trained image encoder provided by Sound2Scene~\cite{sung2023sound} to extract the ground truth audio and image features.

\noindent\textbf{Image generator (generation module).} Fig.~\ref{fig:model} illustrates the generation module of our design. With our approach, two types of images can be generated by the image generator $G(\cdot)$. The types of image generation are categorised as `audio-visual generation' and `audio-visual separation' as shown in Fig.~\ref{fig:model} by red and blue arrows, respectively. Details of each task are as follow:

\begin{itemize}[leftmargin=*]
\item\textbf{Audio-visual generation:}
    We aim to generate a single image containing all the classes in the end. This can be done by combining each separated embedding from the audio-visual separator. When the combined embedding with the control parameter $\lambda$ is given to the image generator, it produces images containing classes corresponding to each sound. We refer to this process as `audio-visual generation', as it is an audio-visual generation task. 
    \item \textbf{Audio-visual separation:}
    We propose a new task in audio-visual separation, which is the second task that our design can achieve. Our proposed audio-visual separation task aims to generate images for each separated audio source. For example, when the `bird and stream' sound is given to the model, the model generates two separate images (i.e., a `bird' image and a `stream' image). This can be achieved by generating each image using each embedding separated by our audio-visual separator.
\end{itemize}

We use a pre-trained image generator provided by the baseline~\cite{sung2023sound}. The pre-trained image generator uses BigGAN~\cite{Brock2018LargeSG-Biggan} with the conditional approach of ICGAN~\cite{casanova2021instanceconditioned-icgan}. 

\noindent\textbf{Mathematical description of our model.}
As we adopt the Sound2Scene~\cite{sung2023sound} backbone as our pipeline, we try to keep their notations for reproducibility and adapt some modifications for our task. As described in Section~\ref{section:data}, our audio-visual inputs contain two ground truth images, two ground truth audio and a mixed sound, denoted as $\mathcal{D} = \{V^1_i, V^2_i, A^1_i, A^2_i, A^{mix}_{i}\}_{i=1}^N$, where $V^j_i$, $A^j_i$, $A^{mix}_i$ represents visual (image) $j\in\{1,2\}$, audio $j\in\{1,2\}$ and mixed audio, respectively.
Likewise, the extracted features from the encoders are denoted as follows:
\begin{itemize}[leftmargin=*]
    \item audio features extracted by the audio encoder $f_A(\cdot)$:
    \begin{itemize}[label=$\bullet$, nosep]
      \item $\mathbf{z^{\scriptscriptstyle{A_j}}_{\scriptscriptstyle{GT}}} = f_A(A^j)$: audio features from GT audio j, $\mathbf{z^{\scriptscriptstyle{A_j}}_{\scriptscriptstyle{GT}}} \in {R}^{2048}$;
    \end{itemize}
    \item image features extracted by the image encoder $f_V(\cdot)$:
    \begin{itemize}[label=$\bullet$, nosep]
      \item $\mathbf{z^{\scriptscriptstyle{V_j}}_{\scriptscriptstyle{GT}}} = f_V(V^j)$: visual (image) features from GT audio j, $\mathbf{z^{\scriptscriptstyle{V_j}}_{\scriptscriptstyle{GT}}} \in {R}^{2048}$;
    \end{itemize}
    \item audio-visual features extracted by the audio-visual separator $f_{AV}(\cdot)$:
    \begin{itemize}[label=$\bullet$, nosep]
      \item $\mathbf{z^{\scriptscriptstyle{AV_{mix}}}} = f_{AV}(A^{mix})$ audio-visual features from mixed audio, $\mathbf{z^{\scriptscriptstyle{AV_{mix}}}} \in {R}^{4096}$;
      \item $\mathbf{z^{\scriptscriptstyle{{AV}_{1}}}}$: separated audio-visual features 1, $\mathbf{z^{\scriptscriptstyle{AV_{1}}}} \in {R}^{2048}$;
      \item $\mathbf{z^{\scriptscriptstyle{AV_{2}}}}$: separated audio-visual features 2, $\mathbf{z^{\scriptscriptstyle{AV_{2}}}} \in {R}^{2048}$.
    \end{itemize}
\end{itemize}

\subsection{Training and Inference}
In this subsection, we show how we train our model and obtain images from the model.

\subsubsection{Training}
\noindent\textbf{Training procedure.} We train an audio-visual separator, which separates audio sources into distinct audio embeddings that are aligned with ground truth audio and image embeddings.
Our model extracts audio features as an audio embedding from a mixed audio input containing two audio sources. The first and the second half of the embedding correspond to each of the classes present in mixed audio, respectively. Each is compared with a ground truth audio embedding and aligned with ground truth image embeddings. Specifically, each split audio embedding is first compared with an audio embedding extracted from unmixed, single-source audio serving as the ground truth. This comparison embedding is extracted using the pre-trained Sound2Scene~\cite{sung2023sound} audio encoder. This step places each of our audio embeddings near the ground truth single audio embedding. Then, the audio embedding is compared to the image embedding to align and place the separated audio embedding with the image embedding. We refer to these steps as A2A alignment (Audio-to-Audio alignment) and A2V alignment (Audio-to-Visual alignment), respectively. 

For A2A alignment, we use InfoNCE losses, and labels are not used for training due to the contrastive learning method. As described in Sound2Scene~\cite{sung2023sound}, InfoNCE is written as: 
\begin{equation}
\label{eq:infonce}
\text{InfoNCE}(\mathbf{a}_j, \{\mathbf{b}\}_{k=1}^{N}) = -\log \frac{\exp\left(-\|\mathbf{a}_j - \mathbf{b}_j\|_2\right)}{\sum_{k=1}^{N} \exp\left(-\|\mathbf{a}_j - \mathbf{b}_k\|_2\right)},
\end{equation}
where \(\mathbf{a}\) and \(\mathbf{b}\) denote arbitrary vectors in the feature space with the same dimension~\cite{sung2023sound}, \(j\) denotes the index of the positive sample among the batch, \(N\) denotes the total number of samples in the batch. The InfoNCE for our A2A alignment is expressed as follows:
\begin{equation}
L_{\text{InfoNCE}} = \frac{1}{2B} \sum_{j=1}^{B} \left( L^{AV}_j + L^A_j \right),
\end{equation}
where our audio-visual-centric loss $L^{AV}_j = \text{InfoNCE}(\mathbf{\hat{\mathbf{z}}^{AV}_j}, \{\hat{\mathbf{z}}^\mathbf{A}\})$ and audio-centric loss $L^{A}_j = \text{InfoNCE}(\mathbf{\hat{\mathbf{z}}^{A}_j}, \{\mathbf{\hat{\mathbf{z}}^{AV}}\})$~\cite{sung2023sound}. $\mathbf{\hat{\mathbf{z}}^{AV}}$ and $\mathbf{\hat{\mathbf{z}}^{A}}$ are the unit vector of the extracted feature from our audio-visual separator and Sound2Scene audio encoder, respectively. 

For A2V alignment, we choose to use InfoNCE loss, as it has been found to be an effective way to align these two different modalities~\cite{sung2023sound}. To express the InfoNCE loss for the A2V alignment, we describe the equation as follows: 
\begin{equation}
L_{\text{InfoNCE}} = \frac{1}{2B} \sum_{j=1}^{B} \left( L^{AV}_j + L^V_j \right),
\end{equation}
where $\mathbf{\hat{\mathbf{z}}^{AV}}$ and $\mathbf{\hat{\mathbf{z}}^{V}}$ are the unit vector of the extracted feature from our audio-visual separator and Sound2Scene audio encoder, respectively. Our audio-visual-centric loss $L^{AV}_j = \text{InfoNCE}(\mathbf{\hat{\mathbf{z}}^{AV}_j}, \{\mathbf{\hat{\mathbf{z}}^V\}})$ and visual-centric loss $L^{V}_j = \text{InfoNCE}(\mathbf{\hat{\mathbf{z}}^{V}_j}, \{\mathbf{\hat{\mathbf{z}}^{AV}\}})$~\cite{sung2023sound}.

\subsubsection{Inference}

With the trained audio-visual separator, images are generated by an image generator. Using the separated embeddings from the audio-visual separator, two types of images can be inferred by the audio image generator. First, unmixed separated images can be generated, where each image contains the class corresponding to one of the separated audio embeddings, allowing for two distinct images to be generated. In addition, if these two embeddings are combined, the image generator produces an image that includes all classes present in the mixed audio input. For the image generator, we use a BigGAN-based architecture~\cite{Brock2018LargeSG-Biggan} with the ICGAN~\cite{casanova2021instanceconditioned-icgan} method, pre-trained by the authors of the baseline~\cite{sung2023sound} in order to obtain a fair comparison of results in Sound2Scene. 

The following equations show how images are generated using the pre-trained image generator 
$G$~\cite{sung2023sound}.
\begin{equation}
G(\mathbf{z^{\scriptscriptstyle{N}}}, \lambda \mathbf{z^{\scriptscriptstyle{AV_{1}}}}+(1-\lambda)\mathbf{z^{\scriptscriptstyle{AV_{2}}}})\label{eq:lambda}\end{equation}
where \( \mathbf{z^{\scriptscriptstyle{N}}} \sim \mathcal{N}(0, I) \), \(\mathbf{z^{\scriptscriptstyle{AV_{1}}}}\), \(\mathbf{z^{\scriptscriptstyle{AV_{2}}}}\) and $\lambda$ denotes noise, separated audio-visual feature 1, separated audio-visual feature 2 from our audio-visual separator and the control parameter, respectively. To combine the two embeddings and generate mixed images, we simply add the embeddings of the same dimensionality with the control parameter $\lambda$ as shown in Eq.~\ref{eq:lambda}\footnote{Images in figures are generated using $\lambda=0.5$.}. It is noteworthy that there are other possible methods that could be applied to image generation, as this is a separate research area.

\section{Results}

\label{sec:results}

\subsection{Experimental Setup}

\noindent\textbf{Dataset.} VGGSound is an audio-visual dataset consisting of 10 seconds of YouTube video clips and the corresponding class labels~\cite{Chen20VggSound}. The dataset consists of 300 classes and 210,000 videos. Since this dataset provides short video clips, it is useful as audio and visuals can be extracted. For our experiment, we first extract the audio from 10-second video clips and extract a frame from each video. When frames are extracted from the video, we use `the highly-correlated frame numbers' provided by the Sound2Scene authors~\cite{sung2023sound} to ensure the relevance of the frame with the audio. We randomly selected 20,000 input samples from 20 combinations of audio classes in VGGSound
\footnote{Details are provided in the supplementary materials.} by applying the methodology in Section~\ref{section:data}. Out of 20 sound classes, 15 are `realistic sounds' that can be heard in everyday situations whereas the remaining 5 are `unrealistic sounds' (not naturally occurring). We aligned the first half of the embedding vector to foreground classes and the second half to background classes. Class labels were not used for training.

\noindent\textbf{Evaluation metrics.} 

We use FID, IS, CRS, R@K as evaluation metrics. Even though improving image quality is not the main goal of this work, we provide evaluation results using image quality metrics~(FID, IS) for reference. Note that FID can only be measured in the audio-visual separation task as it requires ground truth images which are not available for `mixed' ground truth images. In particular, we use R@2\textsuperscript{*}\footnote{We use R@2\textsuperscript{*} as a variation of R@2~\cite{sung2023sound} for our task.} and CRS which are suitable for our task~(given in mixed audio).

\begin{itemize}[leftmargin=*, noitemsep]
    \item Fréchet Inception Distance (FID)~\cite{fid}: FID is a metric to measure the quality of generated images. It measures the similarity between ground truth images and generated images by calculating the distance between feature vectors of two kinds of images using Inception-v3~\cite{Szegedy2015RethinkingTIinception}.
    \item Inception Score (IS)~\cite{Salimans2016ImprovedTF-isscore}: IS score is an evaluation metric to measure the diversity and the quality (fidelity) of generated images. A higher score means that the model generates high-quality images and a variety of images. The score is calculated using KL divergence between the conditional label and marginal distribution~\cite{sung2023sound}. To calculate the score, we use TorchMetrics~\cite{Detlefsen2022TorchMetricsM} to compute Inception-v3~\cite{Szegedy2015RethinkingTIinception}. 
     \item Class Representation Score (CRS): We propose \textit{Class Representation Score~(CRS)} as a new metric for the audio-visual generation task, particularly for the generation from mixed audio. CRS metric calculates how many images actually contain all the sound classes using a state-of-the-art open-vocabulary object detection model (YOLO-World~\cite{Cheng2024YOLOWorld} in our experiments) as follows: 
    \begin{equation}
    CRS_N = \frac{\text{\# of  images containing all objects 1...N}}{\text{\# of total images}},
        \label{generalized_sd_cpds}
    \end{equation}
    where the CRS is represented as a percentage.  
    \item R@K~\cite{sung2023sound}: To calculate R@K scores, candidate class labels for audio and generated images are fed into CLIP~\cite{radford2021learning-clip}, and the similarity between the two modalities is measured. The similarities for each candidate label are ranked, and the number of images where the ground truth label is within the top K is calculated~\cite{sung2023sound}. For a mixed audio input, we use R@2\textsuperscript{*}~(a variation of R@2) which calculates whether both classes present in the mixed audio are ranked within the top 2. For the separation results, we use R@1 which checks if the ground truth label is ranked first.

\end{itemize} 

\noindent\textbf{Implementation details.} We use Sound2Scene~\cite{sung2023sound} as our baseline since it shares the same encoders to extract features and the image generator. We set the batch size to 16 and train the model until epoch 50 using Adam optimiser~\cite{Kingma2014AdamAM} with the learning rate of $10^{-3}$ and the weight decay of $10^{-5}$, same as the baseline~\cite{sung2023sound}. The model is selected based on the training and validation loss graphs, choosing the epoch with the lowest validation loss while ensuring the training loss is decreasing to avoid overfitting. We train our model on a single NVIDIA A100 GPU using a HPC~\cite{kings2022create} cluster. For the evaluation of our model, we use YOLOv8x-worldv2~\cite{Cheng2024YOLOWorld} from Ultralytics~\cite{Jocher_Ultralytics_YOLO_2023} to calculate CRS.

\subsection{Audio-Visual Generation (Mixed Image Generation)}

\subsubsection{Qualitative Results}
    \begin{figure*}[t!]
        \centering
        \includegraphics[width=\linewidth]{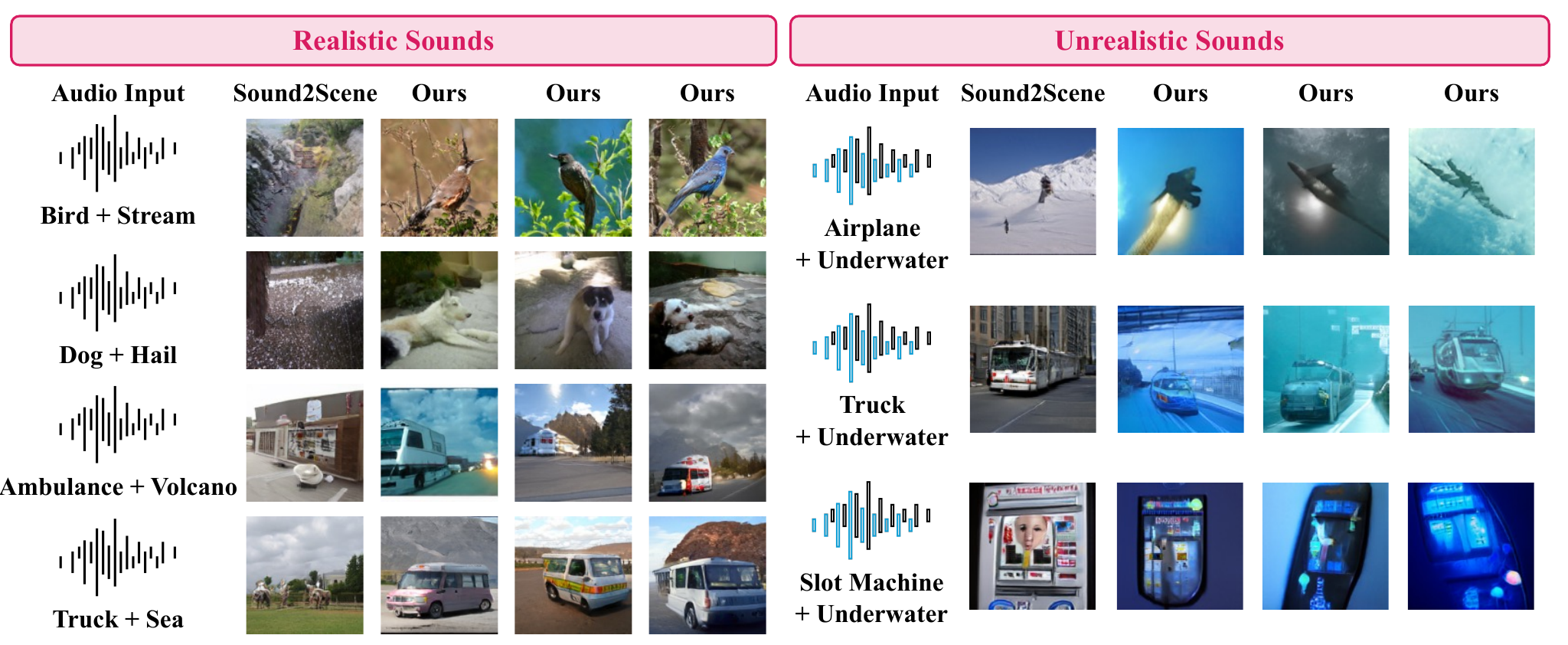}
        \caption{\textbf{Results of audio-visual generation~(generating images given mixed audio)} Existing methods~\cite{av_vae_pedersoli2022estimating(Baseline), av_gan7_fanzeres2022soundtoimagination(S2I), sung2023sound} focus on generating images from a single-class audio input, whereas our method focuses on generating images from a mixed audio input. We show the results from the state-of-the-art~\cite{sung2023sound} and ours based on realistic and unrealistic mixed sounds.  The results given a mixed audio input show how well our model generates images given mixed audio. The first two columns in each part are the results of the state-of-the-art~\cite{sung2023sound} and ours~(\textsc{AV-GAS}) when the same mixed audio is given. The third and fourth columns show other images generated by our approach.}
        \label{fig:result1}
    \end{figure*}

        To compare our approach to existing methods we choose Sound2Scene~\cite{sung2023sound}, which shares a similar goal of image generation but focuses on single-class audio. Even though other methods~\cite{av_vae_pedersoli2022estimating(Baseline), av_gan7_fanzeres2022soundtoimagination(S2I)} have been used as baselines against Sound2Scene, we compare our method only to Sound2Scene, since others are either not designed for audio-visual generation~\cite{av_vae_pedersoli2022estimating(Baseline)} or not publicly available~\cite{av_gan7_fanzeres2022soundtoimagination(S2I)}. 
        
Fig.~\ref{fig:result1} shows that the state-of-the-art model~\cite{sung2023sound} struggles with a mixed-audio input. When mixed audio is given to the Sound2Scene~\cite{sung2023sound}, the model fails to generate images including all classes present in audio, whereas our approach succeeds in representing both classes in the generated images.

It is also noteworthy that our method can `imagine' unrealistic situations by listening to the unrealistic sounds. Even though our possible everyday foreground objects are mixed with underwater sounds, our model can generate images by imagining how the scenes might look.

\subsubsection{Quantitative Results}

Table~\ref{table:mixed} shows the comparison of mixed image generation results between ours and the state-of-the-art~\cite{sung2023sound}. We experiment with various alignment methods (A2A, A2V, A2A+A2V) using InfoNCE. 

As shown in Table~\ref{table:mixed}, CRS and R@2\textsuperscript{*} are significantly higher in our method (A2A). Our model achieves approximately 7\% higher CRS and 4\% higher R@2\textsuperscript{*} than the state-of-the-art~\cite{sung2023sound}. This indicates that our method more successfully generates images containing both foreground and background compared to Sound2Scene: our model significantly outperforms the state-of-the-art model in generating images that include all audio classes. Note that CRS and R@2\textsuperscript{*} are the evaluation metrics that best align with our goal of generating images representing multiple classes present in an audio signal. 

 We also provide the FID and IS results, which focus on measuring image quality rather than the presence of all classes, to give a reference for the quality of generated images even though this is not our main purpose. The main difference between FID and IS is that FID requires ground truth images whereas IS does not. Accordingly, we evaluate only the Inception Score (IS) for our audio-visual generation task because mixed ground truth images are not included in our training and test datasets. Notably, the Inception Score of our method is similar to, or even higher than that of Sound2Scene even though it is trained with additional constraints of modality alignment.

\begin{table}[h!]
\centering
\resizebox{\linewidth}{!}{
\begin{tabular}{l@{\extracolsep{2pt}}c@{\extracolsep{2pt}}c@{\extracolsep{2pt}}c@{\extracolsep{-1pt}}c@{\extracolsep{-1pt}}c@{\extracolsep{5pt}}c@{\extracolsep{5pt}}c@{\extracolsep{5pt}}c}
\toprule
\multirow{2.5}{*}{Methods} & \multicolumn{4}{c}{Alignments}  & \multicolumn{4}{c}{Evaluation Metrics}\\ 
\cmidrule(lr){2-5}  \cmidrule(lr){6-9}
 &  &  & {A2A\textsuperscript{†}} & {A2A\textsuperscript{†}}& FID ($\downarrow$) & IS ($\uparrow$) & CRS (\%) ($\uparrow$) & R@2\textsuperscript{*} (\%) ($\uparrow$) \\ 

\midrule
Sound2Scene \textsuperscript{1}       &   &       &      &    & - & 4.18 $\pm$ 0.06  &  0.04 $\pm$ 0.05 & 0.02 $\pm$ 0.00 \\ 
\midrule
\textbf{Ours (A2A)}   &   &  &  \checkmark  &      & - & 3.71 $\pm$ 0.03 & \textbf{0.11 $\pm$ 0.12} & \textbf{0.06 $\pm$ 0.06}\\ 
Ours (A2V)    &   &   &      &    \checkmark  & - & \underline{4.39 $\pm$ 0.04} & 0.07 $\pm$ 0.11 & \underline{0.02 $\pm$ 0.02}\\ 
Ours (A2A+A2V) &   &      & \checkmark  &  \checkmark & - & \textbf{4.47 $\pm$ 0.05} & \underline{0.09 $\pm$ 0.11} & 0.00 $\pm$ 0.03 \\ 
\bottomrule
\end{tabular}}

\vspace{1pt} 
{\raggedright\fontsize{5.8}{8}\selectfont  \parbox{\linewidth}{† A2A: audio-to-audio alignment, A2V: audio-to-visual alignment}\vspace{-4pt}}
{\raggedright\scriptsize  \parbox{\linewidth}{\textsuperscript{1} Results of single image generation given a mixed audio input.\vspace{0pt}}}
\caption{\textbf{Ablation study of methods (mixed images).} Our approach~(A2A) outperforms the state-of-the-art~\cite{sung2023sound}, achieving 7\% higher CRS and 4\% higher R@2\textsuperscript{*} in generating images that include all classes given a mixed audio input. The best-performing methods are shown in bold, and the second best-performing methods are underlined.}
\label{table:mixed}
\end{table}

\subsection{Audio-Visual Separation (Separated Image Generation)}

\subsubsection{Qualitative Results}

\begin{figure}[t!]
        \centering
        \includegraphics[width=\linewidth]{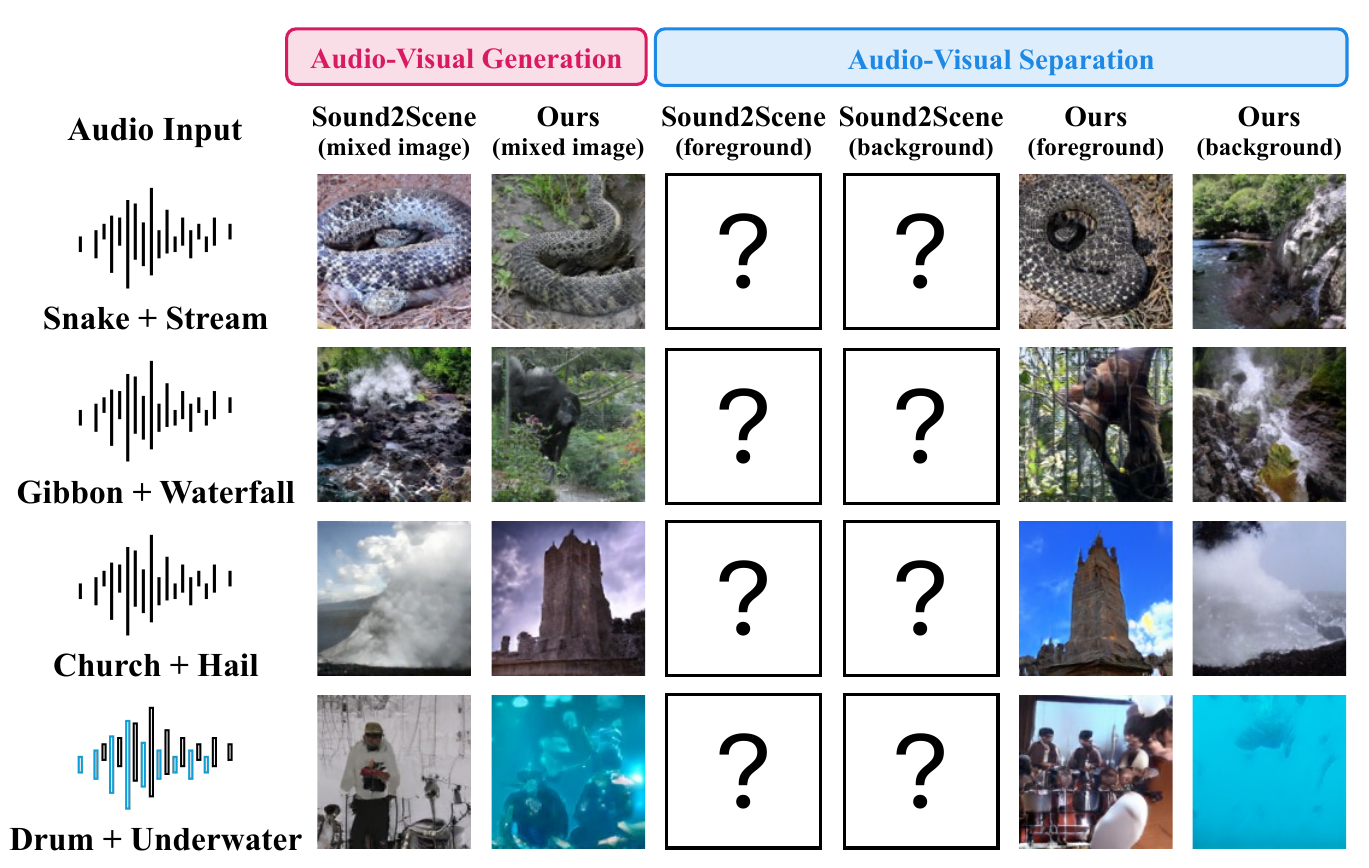}
        \caption{\textbf{Two types of images generated from our model.} Our model generates two types of images: mixed images~(task: audio-visual generation, highlighted in red) and separated images~(task: audio-visual separation, highlighted in blue). Each row shows when the same mixed audio is given to our model and the state-of-the-art~\cite{sung2023sound}. The existing method fails to plausibly generate images given mixed audio, whereas our approach generates images containing all sound classes. Moreover, our model can generate a separate image for each audio class which is not feasible in previous approaches.}
        \label{fig:result2}
    \end{figure}

The results of mixed and separated images are illustrated in Fig.~\ref{fig:result2}. For comparison, the results of Sound2Scene~\cite{sung2023sound} are shown when mixed sound is given as input. Each column represents results from Sound2Scene~\cite{sung2023sound} and our method generating mixed images (task: audio-visual generation), and generating separated images (task: audio-visual separation). Each row represents the results of representative classes we use for the experiments.

As clearly seen from Fig.~\ref{fig:result2}, Sound2Scene~\cite{sung2023sound} fails to generate images containing both classes present in audio. It tends to show either foreground~(e.g., `snake') or background~(e.g., `waterfall') instead of combining both of them. However, our method successfully generates images that include both of them~(audio-visual generation). In addition, our proposed audio-visual separator enables the generation of separated images from mixed audio~(audio-visual separation) which is not achievable in Sound2Scene.

\subsubsection{Quantitative Results}

\noindent\textbf{Comparison on separated images (foreground).}
Table~\ref{table:sep1} shows the comparison of separated object images. Since Sound2Scene does not contain the capability of sound separation, it cannot generate separated foreground images from mixed audio. In contrast, the results of \textsc{AV-GAS} (ours) are the outputs when the model is given mixed audio to generate separated foreground images~(task: audio-visual separation).

\begin{table}[h!]
\centering
\resizebox{\linewidth}{!}{%
\begin{tabular}{l@{\extracolsep{2pt}}c@{\extracolsep{2pt}}c@{\extracolsep{2pt}}c@{\extracolsep{-1pt}}c@{\extracolsep{-1pt}}c@{\extracolsep{5pt}}c@{\extracolsep{5pt}}c@{\extracolsep{5pt}}c}
\toprule
\multirow{2.5}{*}{Methods} & \multicolumn{4}{c}{Alignments}  & \multicolumn{4}{c}{Evaluation Metrics}\\ 
\cmidrule(lr){2-5}  \cmidrule(lr){6-9}
 &  &  & {A2A\textsuperscript{†}} & {A2A\textsuperscript{†}}& FID ($\downarrow$) & IS ($\uparrow$) & CRS (\%) ($\uparrow$) & R@1 (\%) ($\uparrow$) \\ 

\midrule
Sound2Scene \textsuperscript{1} &   &       &      &    & - & -  &  - & -\\ 

\midrule

\textbf{Ours (A2A)}   &   &  &   \checkmark   &      & \textbf{220.26 $\pm$ 31.30} & \textbf{4.48 $\pm$ 0.03} & \textbf{0.47 $\pm$ 0.24} & \textbf{0.47 $\pm$ 0.21}\\ 
Ours (A2V)    &   &   &      &   \checkmark   & \underline{310.23 $\pm$ 32.22} & 3.52 $\pm$ 0.04 & \underline{0.10 $\pm$ 0.10} & 0.09 $\pm$ 0.09\\ 
Ours (A2A+A2V)   &   &      &   \checkmark   &  \checkmark  & 316.22 $\pm$ 33.96 & \underline{3.66 $\pm$ 0.03} & 0.08 $\pm$ 0.12 & \underline{0.12 $\pm$ 0.16} \\ 

\bottomrule
\end{tabular}}
\vspace{0pt} 
{\raggedright\fontsize{5.8}{8}\selectfont  \parbox{\linewidth}{† A2A: audio-to-audio alignment, A2V: audio-to-visual alignment}\vspace{-4pt}}
{\raggedright\scriptsize  \parbox{\linewidth}{\textsuperscript{1} Infeasible for the audio-visual separation task given a mixed audio input.}\vspace{0pt}}
\caption{\textbf{Ablation study of methods (separated foreground images).} In the task of audio-visual separation with a mixed audio input, our approach~(A2A) for the foreground performs the best, indicating that using only audio-to-audio alignment is most effective. The best-performing methods are shown in bold, and the second best-performing methods are underlined.}
\label{table:sep1}
\end{table}

In separated foreground image generation, ours~(A2A) has significantly higher CRS~(47\%) and R@1~(47\%) scores than our other methods, which indicates that using only audio-to-audio~(A2A) alignment is the most effective method to align separated audio embeddings in the audio-visual latent space to generate separated foreground image. 

We include FID and IS results as reference metrics to assess the quality of generated images. 
IS evaluates the quality of separated foreground images from our model, and our values~(4.48) are higher than the results of Sound2Scene in mixed image generation~(4.18). Our model's FID scores are relatively high, which is due to the use of Inception-v3~\cite{Szegedy2015RethinkingTIinception}, trained on a different dataset, and the use of the midframe of the video as the ground truth image for the test set (even though it does not guarantee the inclusion of the audio class). Nevertheless, the Inception Score indicates that our model generates plausible images in terms of quality. Although FID and IS were not the primary focus of our work, we include these metrics here for reference.

\noindent\textbf{Comparison on separated images (background).}
Table~\ref{table:sep2} shows the comparison on separated background images. The experimental conditions are the same as those used in the previous separated foreground image comparison. 

In generating separated background images, most of our methods have a high CRS of approximately 50\%. Our method~(A2A) has the highest CRS score~(57\%), which indicates that using only audio-to-audio alignment results in the best performance, similar to when generating separated foreground images. Regarding FID and IS, they show similar patterns to the results of separated foreground images. 

\begin{table}[ht]
\centering
\resizebox{\linewidth}{!}{
\begin{tabular}{l@{\extracolsep{2pt}}c@{\extracolsep{2pt}}c@{\extracolsep{2pt}}c@{\extracolsep{-1pt}}c@{\extracolsep{-1pt}}c@{\extracolsep{5pt}}c@{\extracolsep{5pt}}c@{\extracolsep{5pt}}c}
\toprule
\multirow{2.5}{*}{Methods} & \multicolumn{4}{c}{Alignments}  & \multicolumn{4}{c}{Evaluation Metrics}\\ 
\cmidrule(lr){2-5}  \cmidrule(lr){6-9}
 &  &  & {A2A\textsuperscript{†}} & {A2A\textsuperscript{†}}& FID ($\downarrow$) & IS ($\uparrow$) & CRS (\%) ($\uparrow$) & R@1 (\%) ($\uparrow$) \\ 

\midrule
Sound2Scene \textsuperscript{1}       &   &       &      &    & - & -  &  - & -\\

\midrule

\textbf{Ours (A2A)}   &   &  &   \checkmark    &      & \textbf{223.89 $\pm$ 25.51} & 3.28 $\pm$  0.03 & \textbf{0.57 $\pm$  0.32} & \underline{0.37 $\pm$  0.23}\\ 
Ours (A2V)    &   &   &      &    \checkmark   & 276.28 $\pm$ 17.58 & \underline{3.78 $\pm$ 0.05} & 0.46 $\pm$  0.28 & 0.31 $\pm$  0.22\\ 
Ours (A2A+A2V)  &   &      &   \checkmark    &   \checkmark   & \underline{260.92 $\pm$ 27.85} & \textbf{3.80 $\pm$ 0.04} & \underline{0.49 $\pm$  0.32} & \textbf{0.43 $\pm$  0.29}\\ 
\bottomrule
\end{tabular}}

\vspace{0pt} 
{\raggedright\fontsize{5.8}{8}\selectfont  \parbox{\linewidth}{† A2A: audio-to-audio alignment, A2V: audio-to-visual alignment}\vspace{-4pt}}
{\raggedright\scriptsize  \parbox{\linewidth}{\textsuperscript{1} Infeasible for the audio-visual separation task given a mixed audio input.}\vspace{0pt}}

\caption{\textbf{Ablation study of methods (separated background images).} In the task of audio-visual separation with a mixed audio input, our approach~(A2A) achieves the highest CRS score for the background and similar scores for the other metrics. The best-performing methods are shown in bold, and the second best-performing methods are underlined.}

\label{table:sep2}
\end{table}

\section{Conclusion}
\label{sec:conclusion}

    In this work, we proposed an audio-visual generative model called \textsc{AV-GAS} to generate images from a soundscape. Our model aims to generate images based on an audio input with multiple classes since our world is surrounded by multiple audio sources. Our proposed audio-visual separator effectively processes such signals and is able to generate images of the classes both jointly and in separate images. This work therefore also introduces a new audio-visual separation task and an appropriate metric: Class Representation Score~(CRS). Our approach generates plausible images from mixed audio containing target classes, achieving 7\% higher CRS and 4\% higher R@2\textsuperscript{*} compared to the state-of-the-art. We anticipate that this work will open new possibilities for processing mixed audio input in audio-visual generation and contribute to the audio-visual research community. In the future, we plan to extend the model to more than two audio sources for generality, since the current method is limited to background/foreground distinctions. We plan to explore different automated strategies for assigning audio sources to locations in the embedding, similar to what is done in music separation~\cite{chen2023iquery}.

{
    \small
    \bibliographystyle{ieeenat_fullname}
    \bibliography{references}
}

\clearpage
{\onecolumn
\setcounter{page}{1}
\maketitlesupplementary}

\renewcommand{\thesection}{\Alph{section}}
\setcounter{section}{0}

\section{20 Classes for Mixed Audio}
\label{sec:rationale}
We provide details of 20 classes used in our experiment. First, we select five possible background sounds that could be heard in real life from among the 50 classes of VGGSound~\cite{Chen20VggSound} provided by Sound2Scene~\cite{sung2023sound}: `hail', `sea waves', `stream burbling', `volcano explosion' and `waterfall burbling'. We then choose three classes as foreground sounds for each of the five background sounds. We call these 15 classes `realistic sounds' to determine whether our model can imagine real-world scenes by hearing the realistic sounds as shown in Table~\ref{table:15classes}. In addition, we create five different `unrealistic sounds' classes. To make these classes, we choose a background sound called `underwater bubbling' and mix it with five different foreground sounds. The purpose of these additional five classes is to see if our model can imagine unrealistic scenes as well as realistic scenes. Therefore, we create 20 classes in total.

\begin{table}[h]
\centering
\resizebox{\textwidth}{!}{
\begin{tabular}{c|cccccc}
\toprule
\multicolumn{6}{c}{15 Realistic Sounds} \\
\midrule
Background & hail & sea waves & stream burbling & volcano explosion & waterfall burbling \\
\midrule
\multirow{3}{*}{Foreground}   & church bell ringing &  ice cream truck, ice cream van & baltimore oriole calling & ambulance siren & gibbon howling \\ 
& dog barking & people cheering & railroad car, train wagon & sheep bleating & people crowd \\
& train horning & playing harp & snake hissing & wood thrush calling &  singing choir  \\
\bottomrule
\addlinespace[0.5em]
\toprule
\multicolumn{6}{c}{Five Unrealistic Sounds} \\
\midrule
Background & \multicolumn{5}{c}{underwater bubbling} \\
\midrule
Foreground  & airplane flyby &  fire truck siren  & orchestra & playing drum kit & slot machine \\ 
\bottomrule
\end{tabular}}
\caption[15 realistic sounds and five unrealistic sounds classes]{\textbf{15 realistic sounds and five unrealistic sounds classes.} The table shows 20 classes of mixed audio consisting of 15 realistic and five unrealistic sounds. The 15 realistic sound classes include five different background sounds, each combined with three different foreground sounds. The remaining five unrealistic sound classes are mixed with `underwater bubbling' as the background sound and five foreground sounds.}
\label{table:15classes}
\end{table}

\section{Prompts Used for Evaluation}
To calculate our proposed evaluation metric \textit{Class Representation Score~(CRS)}, we use YOLO-World~\cite{Cheng2024YOLOWorld} to detect whether classes are present in the image. 
We used the following rules for selecting the prompt that we provide as input to YOLO-World for each class:
\begin{itemize}
    \item We use a single word as a prompt (e.g., `stream' instead of `stream burbling').
    \item We use singular forms instead of plural as we found that YOLO-World detects singular terms more accurately than plural ones (e.g., `person' instead of `people').
    \item In cases where the model struggles with detection, we use alternative prompts (e.g., `mountain' instead of `volcano').
\end{itemize}
Table~\ref{table:prompts} presents the final prompts used for our evaluation. For a fair comparison and due to the model's limitations in detecting objects accurately, we use the same prompts to measure R@2\textsuperscript{*} and R@1 scores. 

\begin{table}[h]
\centering
\resizebox{\textwidth}{!}{
\begin{tabular}{cc|cc|cc}
\toprule
\multicolumn{6}{c}{Prompts Used for the Evaluation} \\
\midrule
Class & Prompt & Class & Prompt & Class & Prompt \\
\midrule
airplane flyby & airplane & orchestra & orchestra & slot machine  & machine\\
ambulance siren & car & people cheering & person & snake hissing & snake \\
baltimore oriole calling  & bird & people crowd & person & stream burbling & stream\\
church bell ringing  & building & playing drum kit & drum & train horning & train \\
dog barking & dog & playing harp & harp & underwater bubbling & aquarium\\
fire truck siren & truck & railroad car, train wagon & train & volcano explosion & mountain\\
gibbon howling & gibbon & sea waves  & sea & waterfall burbling & stream\\
hail & rain & sheep bleating & sheep & wood thrush calling & bird\\
ice cream truck, ice cream van & truck & singing choir & person &  & \\

\bottomrule
\end{tabular}}
\caption[Prompts used for CRS and R@K evaluation.]{\textbf{Prompts used for CRS and R@K evaluation.} To calculate our proposed Class Representation Score~(CRS) or the modified R@K metric, we examine appropriate prompts for evaluation. We establish some specific rules for selecting prompts given to the open-vocabulary object detection model (e.g., YOLO-World~\cite{Cheng2024YOLOWorld}). These rules include using single and singular words, and alternative terms when detection is challenging.}
\label{table:prompts}
\end{table}

\section{Failure Cases}
\begin{figure}[h!]
        \centering
        \includegraphics[width=.25\linewidth]{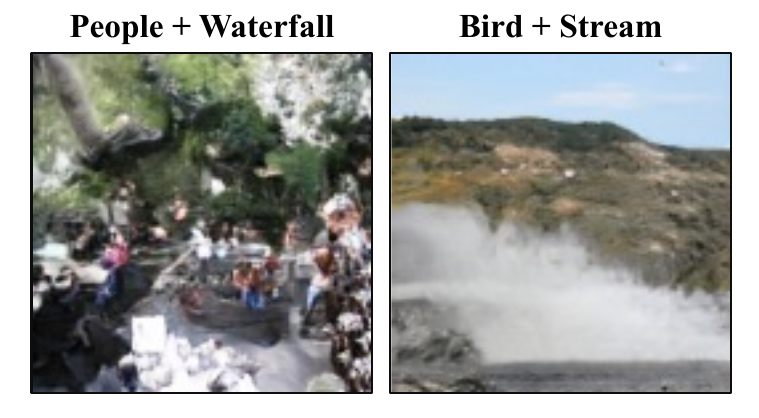}
        \caption{\textbf{Examples of Failure Cases.} The two images show representative failure cases of generated images from our model. The first image corresponds to a generated image from  `people crowd + waterfall burbling' mixed audio where a GAN-based image generator struggles to generate plausible people images. The second image shows when `baltimore oriole calling + stream burbling' mixed audio is given to our model. When the volume of one class in mixed audio ~(e.g., bird sound) is too low, the corresponding object can disappear in the generated images.}
        \label{fig:failure}
    \end{figure}

Fig.~\ref{fig:failure} illustrates failure cases. The GAN-based image generator struggles to generate plausible images involving people, which results in some failures for classes containing people. In addition, when the volume of one class in mixed audio is significantly lower than the other, the corresponding object disappears in the generated images. 

\end{document}